\documentclass{article}
\usepackage{spconf,amsmath,graphicx,pdfpages}
\usepackage{booktabs}

\title{Automatic dataset generation for specific object detection}
\name{Xiaotian Lin, Leiyang Xu, Qiang Wang
\thanks{© 2022 IEEE. Personal use of this material is permitted. Permission from IEEE must be obtained for all other uses, in any current or future media, including reprinting/republishing this material for advertising or promotional purposes, creating new collective works, for resale or redistribution to servers or lists, or reuse of any copyrighted component of this work in other works.}}
\address{Department of Control Science and Engineering, Harbin Institute of Technology, Harbin, China}
\begin{document}

%
\maketitle
\begin{abstract}
In the past decade, object detection tasks are defined mostly by large public datasets. 
However, building object detection datasets is not scalable due to inefficient image collecting and labeling. Furthermore, most labels are still in the form of bounding boxes, which provide much less information than the real human visual system. 
In this paper, we present a method to synthesize object-in-scene images, which can preserve the objects' detailed features without bringing irrelevant information. In brief, given a set of images containing a target object, our algorithm first trains a model to find an approximate center of the object as an anchor, then makes an outline regression to estimate its boundary, and finally blends the object into a new scene. Our result shows that in the synthesized image, the boundaries of objects blend very well with the background. Experiments also show that SOTA segmentation models work well with our synthesized data.
\end{abstract}
\begin{keywords}
Dataset generation, Object detection, Computer vision, Image processing
\end{keywords}
\section{Introduction}

Object detection and recognition is a popular research field and have been widely used in  industry and in daily life.
Because of the breakthrough and rapid adoption of deep learning, many highly accurate object detection algorithms and methods have been proposed. However, one of the biggest drawbacks of using the SOTA object detection system is a large amount of labeled data because the cost of labeling the dataset manually is high. Public datasets, such as PASCAL VOC\cite{Everingham}, MS COCO\cite{lin}, ImageNet\cite{deng} can be used to train a general object detector. However, for a specific object, public datasets may bring the following problems: (1) Public datasets may not contain specific objects. Meanwhile, collecting and labeling images with various circumstances is expensive. (2) Public datasets always assume certain combination between the object and the background. For example, refrigerator and kitchen. This correlation causes the network to extract some features that are not related to the object itself. Such features may increase performance in the competitions on public dataset but will affect the accuracy of object detection when the object migrates to an unknown environment.

One popular way for overcoming these problems is rendering scenes and objects using 3D models \cite{tremblay}\cite{prakash}. Another method is generating a synthetic dataset by placing real segmented object images onto background images. Sagues et al. propose a specialized method to generate a synthetic labeled dataset for kitchen object \cite{sagues}. Georgakis et al. superimpose 2D images of textured object models into images of real indoor environments \cite{georgakis}.
Dwibedi et al. present a method to make detectors ignore the artifacts during training \cite{dwibedi}. \cite{yun} introduces an approach to diversify the foreground with GAN-based seed images in order to balance domain gaps. 
Compared with the existing public dataset, the advantages of synthetic dataset are as follows: (1) The number of synthetic data can be much larger than human labeled data. (2) The objects and backgrounds are not limited to some certain combination. Thus, the model can learn more intrinsic object features. (3) The synthetic dataset can be labeled in an arbitrary way. Contours, masks, bounding boxes and any other forms of annotations related to a practical project are possible.

\begin{figure*}[htb]
  \centering
  \centerline{\includegraphics[width=17cm] {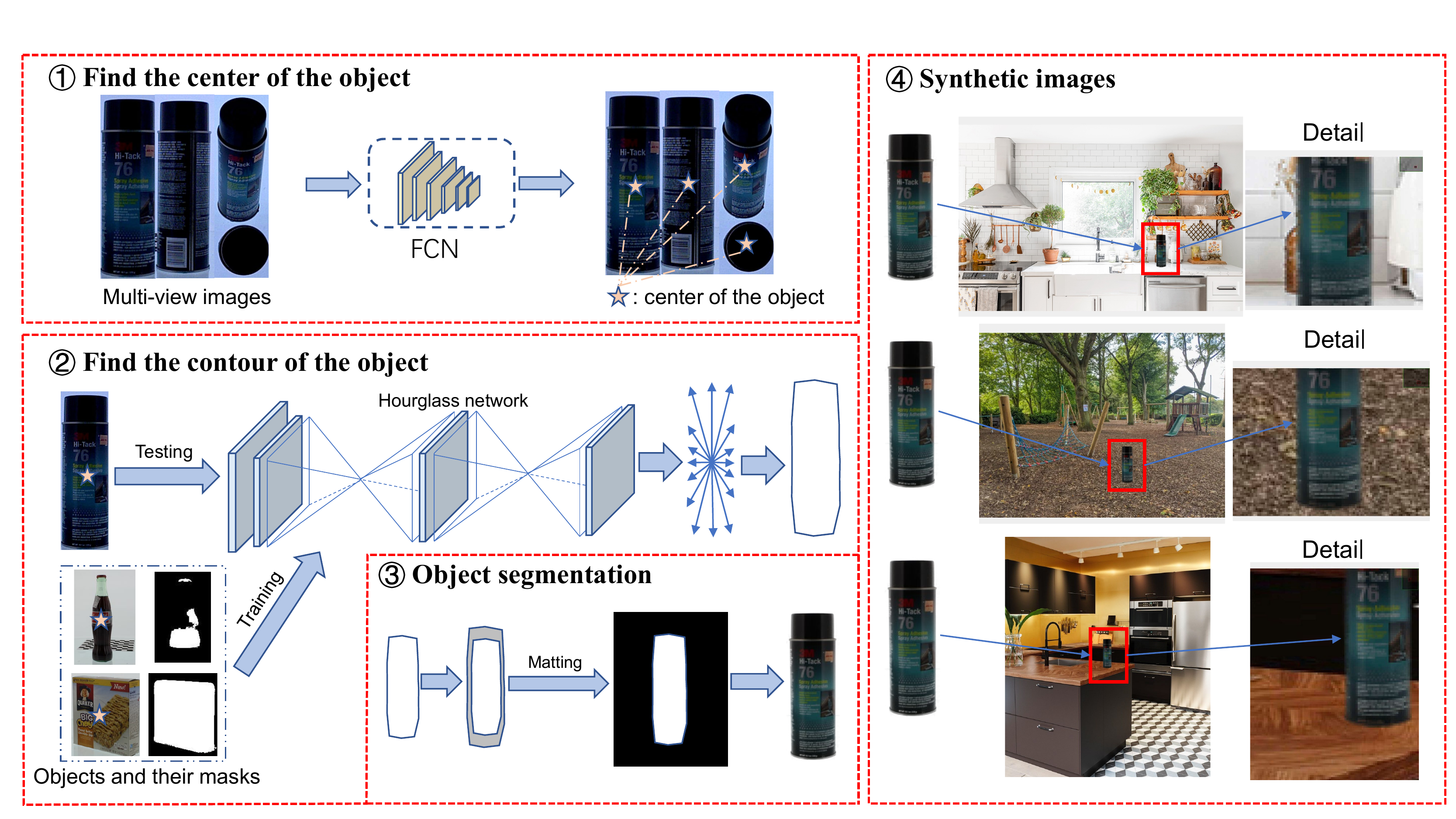}}

  \vspace{-0.3cm}
\caption{Diagram of the proposed process to synthesize images. Some original masks are incomplete as shown in the second part}
\vspace{-0.1cm}
\label{fig:overview}
\end{figure*}

In this paper, we propose a method to generate synthetic dataset. We first get images of different views of the object and apply a convolutional neural network to find the center point of the specific object in the image. Then we train another network to obtain the distance of the object contour relative to the center point  in different directions.
After that, a rough contour of the object is drawn based on these distances. The method in \cite{levin} and \cite{perez} can help us segment the precise contour of the object and blend it into the background. Fig. \ref{fig:overview} briefly shows the whole pipeline of our method.


\section{Method}

\subsection{Object Pool}
Objects are the key components of a synthesized image aimed  at training object detection models. Object images should provide enough diversity of a specific category. More importantly, they shouldn't bring any information of the original background where it came from. Thus, we need an object pool from which object data with detailed features and little noise can be drawn as much as the algorithm needs.

The pool should contain: (1) Raw images with the specific kind of object. (2) High quality binary masks of the corresponding object images. (3) Alpha maps of the corresponding object images. With these data, synthesizing a clean and detailed object-scene image is possible. However, it's not a trivial task to generate such data  with little human interference.  We use the BigBIRD dataset as an example to illustrate the whole pipeline of our algorithm. The BigBIRD dataset provides raw images and some estimated masks. The problem is that these masks is estimated by depth information, many of which is not accurate, even incomplete. With these masks, it's impossible to generate high quality synthesized images because it's hard to tackle with the details on the object boundary. We build a model to find the actual boundaries of these  objects and will show why our model can replace the depth-based mask estimation. The key point of our model is the representation of object contour with a center localization and outline regression, which is similar to Polarmask in \cite{xie_polarmask_2020} and \cite{xie_polarmask_2021}. Comparing with general instance segmentation methods which need to detect instances, our model focuses more on extracting a refined contour of the object.

\subsection{Center Localization}
First, instead of applying popular segmentation models to classify all the pixels into foreground and background, we try to  build a model to find the center of the object.  The reason we do this is that the data we have is not good enough for training a very good segmentation model, which is a relatively hard task and  always needs a lot of fine masks to train on. Also, it turns out that they are not very easy to generalize. But given what we have, it's well enough to train a model whose only task is to regress the center of an object. As we assume there's only one object in raw data, we can build a model based on VGG-16 \cite{simonyan2014very}. The first part is a VGG net pre-trained on PASCAL VOC dataset without the output FC layers, which will extract deep feature maps from the input raw image. As the pre-trained network is already able to classify and localize many objects, the convolutional kernels should be able to extract some common visual features. The parameters in the first part is still learnable. The second part is a small FC layer, which outputs two real numbers. The training input is raw image and the output is two real numbers indicates the x and y bias of ground truth center point normalized by image size. 

Although the mask is not accurate in the aspect of ground truth overlapping, however, as long as the mask is not completely missing, the mass center of the points within the mask will  not be far from the actual mass center of the object. Furthermore, the predicted center location does not need to be as accurate as the boundary we eventually need, so during training, we focus more on the generalizability of the model. 
\subsection{Outline Regression}
We present a special representation of object boundary. We call it outline instead of boundary because we are not building a model to get the actual segmentation of an object, but to find a parameterized boundary representation. As shown in Fig. 2, this form represents the distances between the object center and the object boundary in multiple directions. Specifically, we set up 16 directions evenly distributed around the object center, and the number 16 is a hyperparameter chosen by validation dataset. In recent researches, we can find some similar setups that represent the object as center and polygon. The key difference of our design is that for a certain center point, the degree of freedom is reduced to the number of directions.
\begin{figure}[!t]
  \centering
  \centerline{\includegraphics[width=6cm] {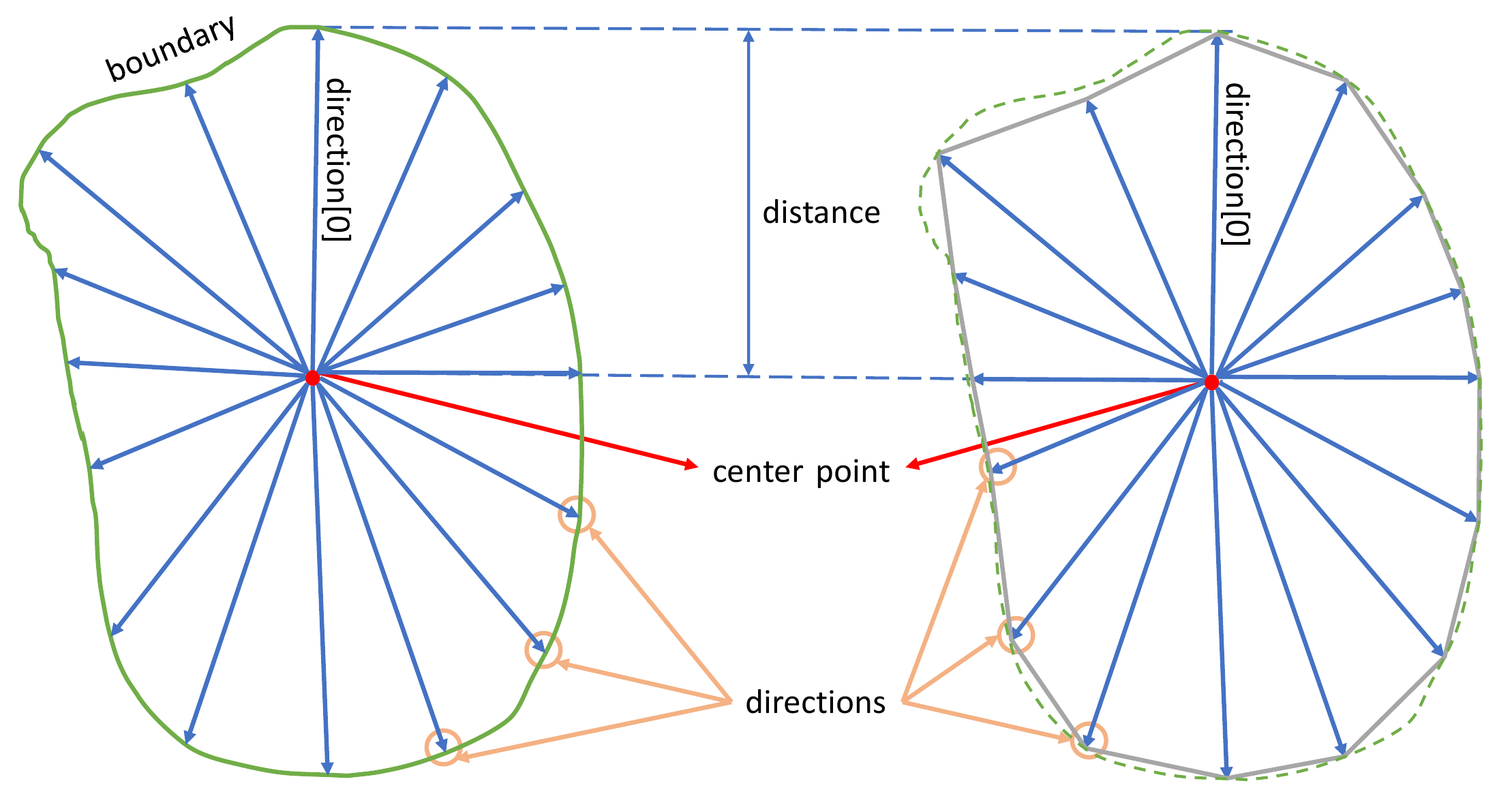}}
\vspace{-0.2cm}
\caption{Illustration of object boundary representation}
\label{fig:outline}
\vspace{-0.2cm}
\end{figure}

There are two main designs in outline regression model. First we need a model to extract features of the actual boundary otherwise the model won't be able to find out the distance between the boundary and a given center point. We choose stacked hourglass \cite{newell} as the backbone feature extractor, because it works well on finding multiscale local features. Hourglass is usually used on finding human body key points and pose estimation. Similarly, the key points on the object boundary is what our model is trying to find. Second, one may notice that the fixed directions make the model sensitive to the predicted center point, that is, if we train the model to find the distances between the boundaries and the actual center point, an error of center point prediction will influence the performance a lot. Our solution is that we make the predicted center point as another input of the outline regression model. We call this input point 'anchor point' (nothing to do with anchor box in YOLO or SSD).
Intuitively, we can treat the anchor point as a new center and make the model output the new distances with respect to the new center. Thus, we need the model to output different distances for the same input image with different anchor points. Furthermore, we need the model to be trained end-to-end, so we should inject this point as a part of the model. Specifically, we add a convolutional layer which makes a translation transformation according to the offset of anchor point with respect to the image center. In contrast with the kernels learned from data, the behavior of this convolutional layer is clear and intuitive.

These two designs are intended to reduce the complexity of our model by adding constraints. The first one transforms the segmentation problem to a reduced number of parameter regression, while the second adds a fixed convolutional kernel which doesn't rely on training. The training procedure is similar to center localization, in which we use pre-trained Hourglass as a feature extractor and use mean square error loss to regress distances output. The different part is how we prepare training data. Based on the above model description, we intentionally set random offsets to the ground truth center point to generate fake anchor points, as well as the distances associated with the generated anchor points. Our model outperforms mask-based segmentation models and the depth-based estimation because it doesn't overfit to some wrong labels in training set as shown in Fig. \ref{fig:overview}.

\subsection{Image Synthesizing}
\textbf{Scene Collection.} After obtaining the object, we need to collect variety of backgrounds. Places365-Standard \cite{zhou2017} is the dataset used for scene recognition. There are 18 million train images from 365 scene categories in this dataset. We select the backgrounds of the synthesized image from this dataset. In this way, the diversity of the synthetic image background  is guaranteed.
 
\begin{figure}[!t]
  \centering
  \centerline{\includegraphics[width=8.5cm] {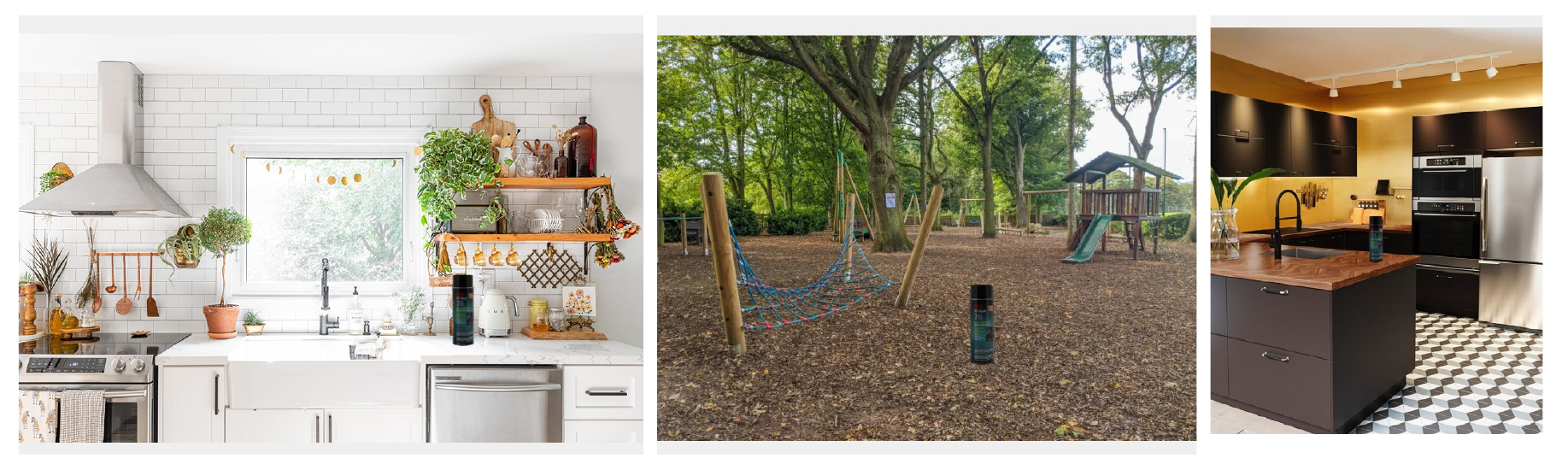}}
\vspace{-0.2cm}
\caption{Examples of the synthetic images}
\label{fig:examples}
\vspace{-0.2cm}
\end{figure}

\textbf{Matting and blending.} In order to make the composite picture look more real, we need to process the  edge of the foreground and background. With the methods we present, a set of boundary key points can be generated. Our solution is to use some traditional image processing methods such as morphology operations to make a more confident but smaller foreground prediction and larger background prediction, with some unknown area in between. These areas can combine as a trimap, which can later be the basic clue of image matting. In our project, we use the closed form image matting to get a relatively high quality mask of the objects and cut it from the original background.
Then we implement Poisson blending to eventually make a synthetic image. Poisson blending  \cite{perez} is an image processing operator that allows the user to insert one image into another, without introducing any visually unappealing seams. Moreover, this method also makes sure that the color of the inserted image is also shifted, so that the inserted object feels as if it is part of the background of the target image. We choose this method because it has closed form solution, which is stable and fast. Some examples of the synthetic images are shown in Fig. 3.

\begin{table}[htbp]
  \centering
  \caption{Evaluation results}
  \vspace{0.1cm}
  \scalebox{0.7}{
  \begin{tabular}{cc}
    \toprule
    Category&Segmentation AP \\
    \midrule
    quaker chewy low fat chocolate chunk&97.062 \\
    white rain sensations apple blossom hydrating body wash&97.909 \\
    suave sweet guava nectar body wash&95.011 \\
    white rain sensations ocean mist hydrating conditioner&90.259 \\
    clif zbar chocolate brownie&97.773 \\
    nature valley granola thins dark chocolate&90.732 \\
    haagen dazs cookie dough&96.465 \\
    dove beauty cream barh&87.339 \\
    honey bunches of oats with almonds&96.822 \\
    spongebob squarepants fruit snaks&96.293 \\
    \midrule
    mean AP over all categories&94.556\\
    \bottomrule    
  \end{tabular}}
\label{table:res}
\vspace{-0.4cm}
\end{table}

\section{Experiments and Discussion}
To verify the data synthesizing pipeline, we test it by applying object detection models. Briefly, the steps are: (1)Randomly draw scene metadata from scene collection, which is used to retrieve an actual image later. (2)For each scene, draw different object metadata to make scene-object pairs. (3)Apply our synthesizing algorithm and generate multiple synthesized images  and labels respectively for each random pair, with respective outline annotation. (4)With the generated images and labels, it is possible to train an object detection model. We show different kinds of detection model to show the flexibility of our synthesizing pipeline. Either with anchor or anchor-free, with bounding box or bounding box free works well with our method.

\subsection{Generate COCO-style dataset}
Generally, bounding box is efficient for building simple models, but it is actually a loss of information. In our work, no human interference is needed for synthesizing a new image, which means that arbitrarily large size of detailed annotations can be generated, with only a slight increase in resource consuming. The synthesized data are labeled with boundary points. As we know, this form of data labeling is very time-consuming and needs a lot of labor, while our data is purely generated by algorithm, fast and cheap.

To show that our algorithm generates rational annotations, we will train an instance segmentation model with the synthesized images and annotations. That's why we generate the annotations in COCO style. With the help of COCO API, it's efficient to train such a model and visualize the results.

\subsection{Train a model with synthesized data}
\begin{figure}[tb]
  \centering
  \centerline{\includegraphics[width=8.5cm] {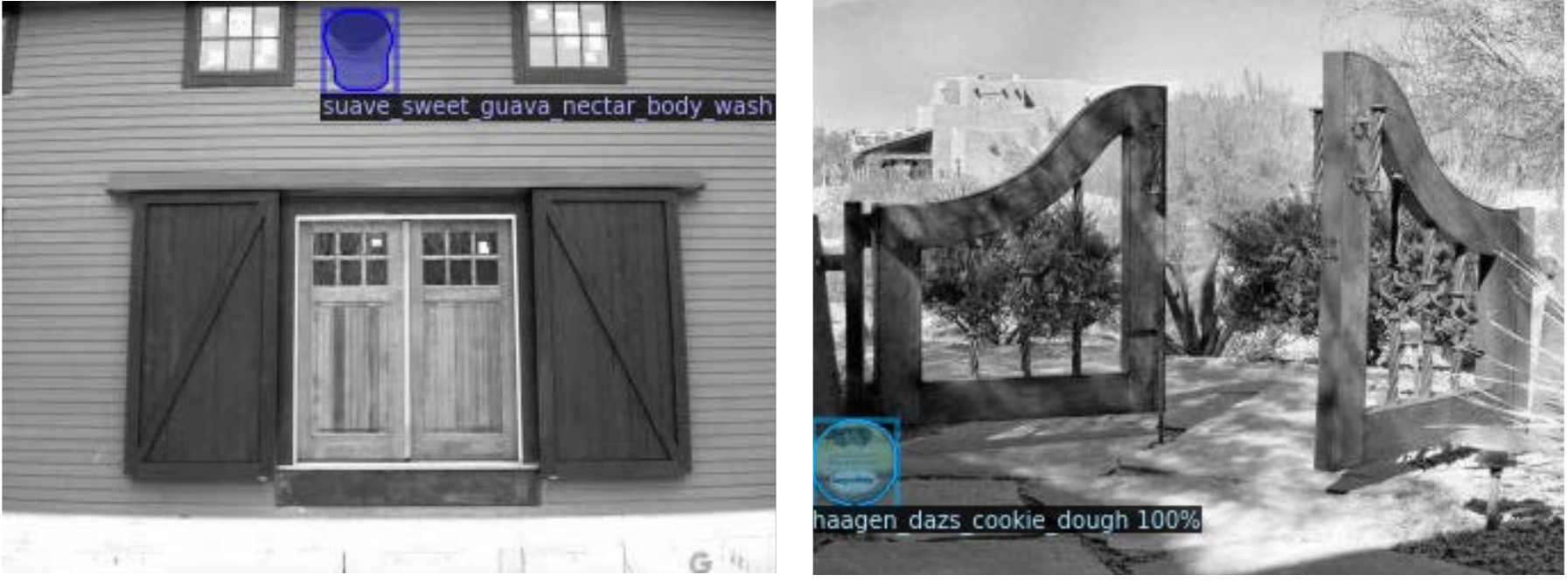}}
\caption{Predictions on unseen data}
\label{fig:validation}\
\vspace{-0.8cm}
\end{figure}

We trained an instance segmentation model based on Mask R-CNN structure. The backbone network is ResNet-50 with FPN neck. The model is originally trained on COCO dataset, which has 80 categories of daily objects. These categories don't cover the objects in BigBIRD dataset. Therefore, we should modify the head of this network to adapt to our new data. For illustration, we only chose 10 random objects and 365 different scenes to synthesized about 100k images with bounding box and mask annotations. The algorithm is fast enough to be finished in only about 1 hour on an Nvidia 3090 GPU. We trained the model on 80\% of the synthesized data and evaluated on the rest. Fig. \ref{fig:validation} shows some results of segmentation during evaluation and Table \ref{table:res} shows the performance for each separate category.

\subsection{Discussion}
Normally, the mean average precision for segmentation on COCO dataset is around 50 percent in recent researches. Table \ref{table:res} shows that our synthesized data is much easier in case of instance segmentation because the number of category is much smaller, the annotations are much more precise and the diversity of a single category is much richer. The synthesized data is not comparable to public datasets as it doesn't help train a general object detector like COCO dataset does. However, it can be applied to a variety of scenarios and the following shows what we are working on.

\textbf{(1) Special objects detection} For an uncommon target, it is generally difficult to find it in public datasets. A conventional solution is to collect many real images which contains this target, label them pixel by pixel and apply transfer learning. While our method only needs some multi-view two-dimensional images of the object and the corresponding rough marks generated by some unsupervised image processing (mask can be incomplete).
\textbf{(2) Testing  existing network} If we have a trained network, and we want to test its performance in specific objects or environment. It is easy for us to generate the synthetic images that meet the requirements.
\textbf{(3) Finding good visual representation} Network structures are usually heuristically designed, which is unlikely to fit all practical circumstances. With arbitrarily enough data of a single object, models based on reinforcement learning or adversarial learning which are always hard to train can be carried out. Then it's possible to learn a more robust visual representation for a specific object under a particular circumstance such as strong noise, limited illumination, frequent occlusion and their combinations.

\section{Conclusion}

In this paper, we proposed a method to generate synthetic dataset. A model was developed to find the outline of the object and segment it from the original target image. A new blending method based on morphology and closed-form matting was proposed. We gave some examples of our synthesized images with scoped details and trained an instance segmentation model on the data generated by our method to show the performance. Finally, we discussed how our synthesized data differs from public datasets and how the proposed method can help us in practical projects.


\bibliographystyle{IEEEbib}
\bibliography{refs}

\end{document}